\documentclass[runningheads]{llncs}
\usepackage[T1]{fontenc}
\usepackage{graphicx}
\usepackage{tikz}
\usepackage{amsmath}
\usepackage{booktabs}
\usepackage[most]{tcolorbox}
\usepackage{xcolor}
\usepackage{mdframed}

\begin{document}
\title{AI Co-Artist: A LLM-Powered Framework for Interactive GLSL Shader Animation Evolution}
\titlerunning{AI Co-Artist: An LLM-Powered System for Interactive Shader Evolution}
%
\author{Kamer Ali Yuksel \and
Hassan Sawaf}
%
\authorrunning{Yuksel et al.}
%
\institute{aiXplain Inc., San Jose, CA, USA\\
\email{\{kamer,hassan\}@aixplain.com}}
%
\index{Yuksel, Kamer Ali}
\index{Sawaf, Hassan}

\maketitle              
\begin{abstract}
Creative coding and real-time shader programming are at the forefront of interactive digital art, enabling artists, designers, and enthusiasts to produce mesmerizing, complex visual effects that respond to real-time stimuli such as sound or user interaction. However, despite the rich potential of tools like GLSL, the steep learning curve and requirement for programming fluency pose substantial barriers for newcomers and even experienced artists who may not have a technical background. In this paper, we present \textit{AI Co-Artist}, a novel interactive system that harnesses the capabilities of large language models (LLMs), specifically GPT-4, to support the iterative evolution and refinement of GLSL shaders through a user-friendly, visually-driven interface. Drawing inspiration from the user-guided evolutionary principles pioneered by the Picbreeder platform, our system empowers users to evolve shader art using intuitive interactions, without needing to write or understand code. AI Co-Artist serves as both a creative companion and a technical assistant, allowing users to explore a vast generative design space of real-time visual art. Through comprehensive evaluations, including structured user studies and qualitative feedback, we demonstrate that AI Co-Artist significantly reduces the technical threshold for shader creation, enhances creative outcomes, and supports a wide range of users in producing professional-quality visual effects. Furthermore, we argue that this paradigm is broadly generalizable. By leveraging the dual strengths of LLMs—semantic understanding and program synthesis—our method can be applied to diverse creative domains, including website layout generation, architectural visualizations, product prototyping, and infographics. We also explore whether human curators in the interactive process could be replaced or augmented with multimodal vision-language models acting as autonomous aesthetic judges to allow closed-loop evolution.
\end{abstract}

\begin{figure*}[t]
  \centering
  \includegraphics[width=\textwidth]{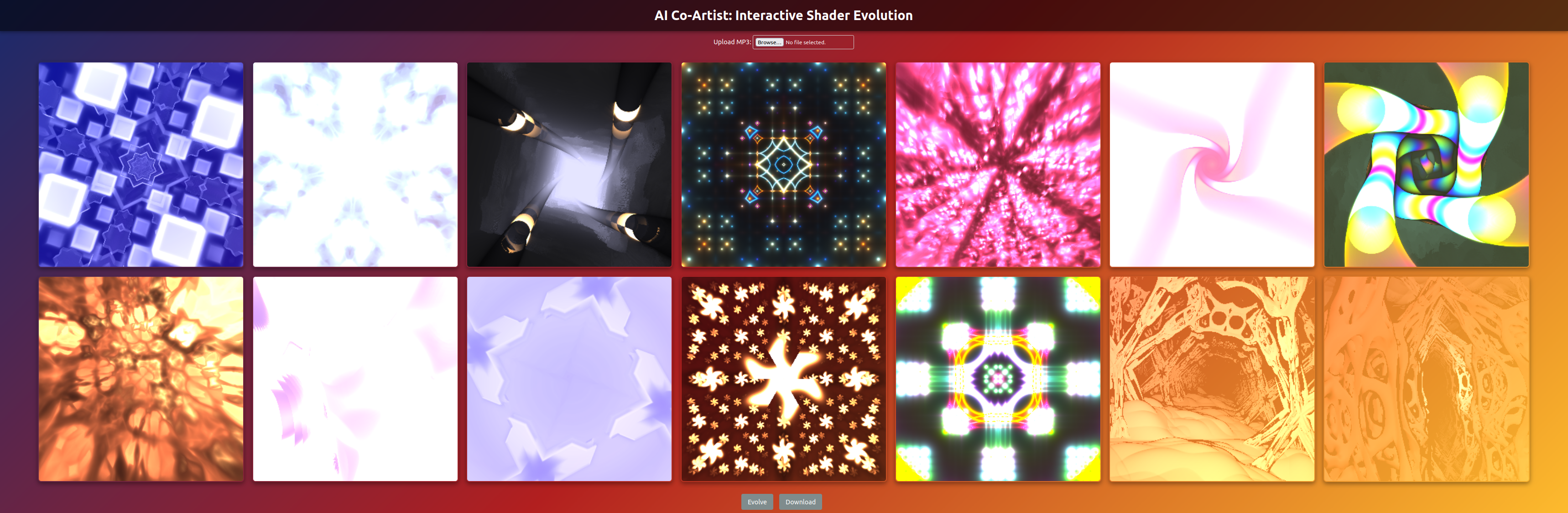}
  \caption{Screenshot of the \textit{AI Co-Artist} interface showing initial audio-reactive shaders, user selection UI, and audio upload controls. Each canvas renders a distinct LLM-generated shader responsive to the uploaded audio file.}
  \label{fig:interface}
\end{figure*}

\begin{figure*}[t]
  \centering
  \includegraphics[width=\textwidth]{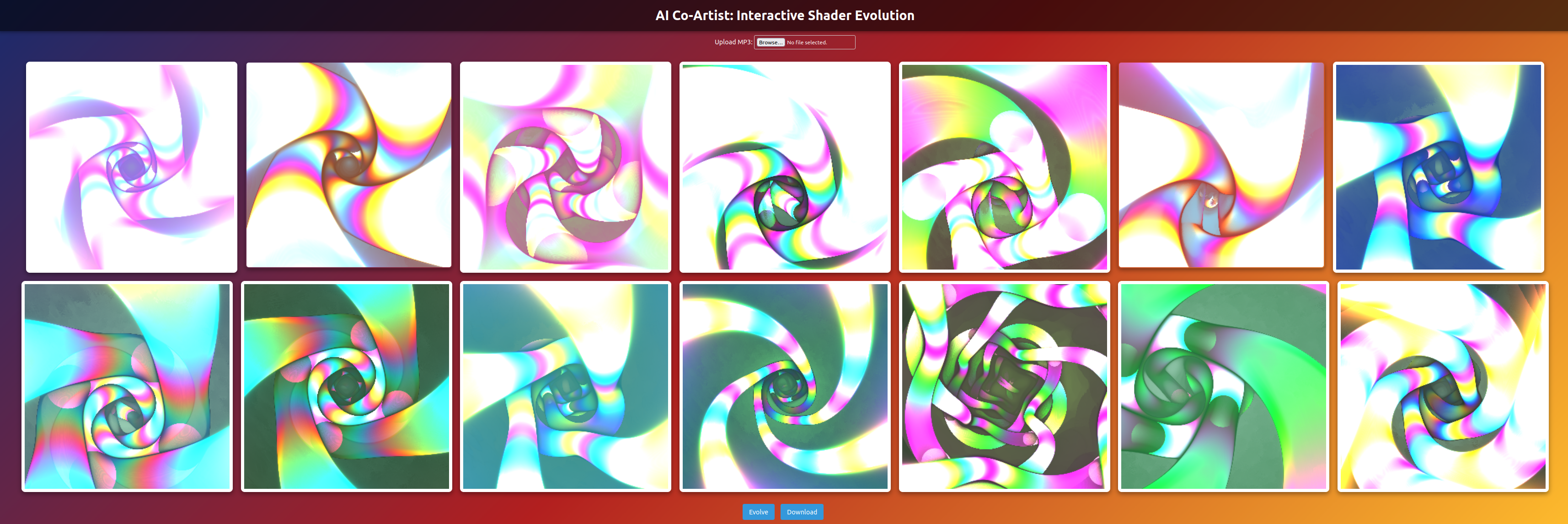}
  \caption{Multiple variants generated from a single parent shader with an interactive evolution. Each square represents a distinct audio-reactive fragment shader generated by the LLM based on user-selected parents. Users view these live-rendered shaders and act as curators by selecting visually compelling ones. The interface encourages the exploration of aesthetic diversity, enabling users to steer the creative direction.}
  \label{fig:shader-variants}
\end{figure*}

\section{Introduction}

The rise of procedural graphics and shader programming has enabled a revolution in digital creativity, particularly in interactive and generative art. Shaders, small programs that run on the GPU, offer fine-grained control over visual outputs at the pixel or vertex level. Yet, this power is often offset by the complexity of the required programming knowledge, which deters many creatives from engaging directly with the medium. Even those with moderate experience in computer graphics may find the trial-and-error nature of shader design challenging, especially when experimenting with reactive or time-varying effects. Parallel to this development, the field of AI-assisted creativity has undergone a transformation driven by LLMs. Trained on massive corpora of natural language and source code, LLMs like GPT-4 have demonstrated remarkable capabilities in tasks such as code synthesis, documentation, transformation, and even creative tasks such as poem writing or art description. Importantly, LLMs are no longer just code completion tools; they are capable of autonomously generating novel, complex programs from vague or incomplete prompts. This unlocks an opportunity for collaborating with AI in creative workflows.

In \textit{AI Co-Artist}, we leverage the generative and interpretive abilities of LLMs to overcome the barriers inherent in shader design. The system positions the user as a creative curator: they select visual outputs they find appealing, and the LLM evolves new shader code based on these preferences. Unlike traditional design tools, the user does not need to specify exact changes or understand the underlying codebase. Instead, they guide the system through aesthetic preference, making the process more accessible, exploratory, and engaging. This interactive evolution process mirrors natural creative cycles—brainstorm, prototype, refine—augmented by an AI assistant capable of producing diverse, structurally valid code variations. The result is a co-design paradigm where both the human and AI contribute meaningfully: the human sets the aesthetic trajectory, while the AI fills in technical details and explores adjacent possibilities. Moreover, because LLMs understand a wide range of domains and programming languages, this framework can be easily extended beyond shaders. It provides a glimpse into a broader future of AI-human collaboration in creative tasks.

This paper presents a fully working implementation of this paradigm in the domain of GLSL fragment shaders. By coupling real-time visual feedback, audio-reactivity, and evolutionary AI-driven code generation, \textit{AI Co-Artist} transforms the shader authoring experience into a playful and intuitive exploration of aesthetic space. The system not only generates functional code—it evolves a creative conversation between the user’s visual preferences and the AI’s expressive potential, realized through structured code transformations like crossover and mutation. A working prototype of the system is available in the supplementary material for experimentation. The HTML can be run locally by anyone interested in exploring the system. To activate the GPT-4-powered shader evolution, users must insert their own OpenAI \texttt{\textbf{\textcolor{red}{API\_KEY}}} in the appropriate variable in the HTML file. The primary contributions of this work can be summarized as:

\begin{itemize}
    \item \textbf{An Interactive Evolutionary Framework for Shader Design:} We introduce a user-guided, LLM-powered system that evolves GLSL fragment shaders through visual selection and automatic code transformation, removing the need for manual programming.

    \item \textbf{LLM-Orchestrated Shader Crossovers and Mutations:} We design prompt-based workflows that allow LLMs to perform crossover and mutation operations on GLSL shaders, yielding valid and visually distinct variants without human intervention.

    \item \textbf{Seamless and Intuitive User Experience:} A web-based interface enables users to browse, select, evolve, and preview shaders effortlessly. Fullscreen previews, progress feedback, and shader download options support creative exploration.

    \item \textbf{A Generalizable Co-Creation Paradigm:} Though focused on shaders, our approach exemplifies a broader class of AI-human creative collaboration tools that can be adapted to other design domains such as music synthesis, generative visuals, or architectural forms.
\end{itemize}

\section{Background and Related Work}

Our work lies at the intersection of interactive evolutionary computation (IEC), AI-assisted code generation, and human-computer creative collaboration. Interactive Evolutionary Computation (IEC) leverages human evaluation to guide evolutionary algorithms, making it particularly effective in domains where quantitative fitness functions are challenging to define, such as art and design. \cite{takagi2001interactive} provides an extensive survey on IEC applications, highlighting its success in addressing complex, real-world problems. A seminal example is \textbf{Picbreeder}, an online platform that enables users to collaboratively evolve images through selective breeding. Users choose images they find visually compelling, which are then mutated and combined to produce new offspring. Over time, this process leads to the emergence of highly structured and aesthetically rich images, demonstrating the power of user-in-the-loop generative systems \cite{secretan2011picbreeder}. Other efforts in evolutionary art have leveraged genetic algorithms for creative generation. For instance, \textbf{GenJam} is an interactive genetic algorithm that evolves jazz solos in collaboration with a human performer \cite{biles1994genjam}. Similarly, the \textbf{NeuroEvolution of Augmenting Topologies (NEAT)} algorithm has been applied to evolve network topologies for various creative tasks \cite{stanley2002evolving}. \cite{lewis2008genetic} discusses the application of genetic programming in evolving artistic designs, emphasizing the role of human aesthetic evaluation in guiding the evolution process.

The advent of Large Language Models (LLMs) has revolutionized code generation and transformation. Models such as OpenAI's \textbf{Codex} have demonstrated remarkable capabilities in generating code snippets from natural language prompts \cite{chen2021codex}. \textbf{GitHub Copilot}, powered by Codex, assists developers by autocompleting code and suggesting alternative implementations, effectively serving as an AI pair programmer. Tools like \textbf{Code Llama} offer AI-driven code completion across multiple programming languages, enhancing developer productivity \cite{wang2023codellama}. Despite their impressive capabilities, these tools are primarily reactive—they extend existing code rather than proactively generate creative alternatives. Our approach differs by prompting the LLM to generate entirely new GLSL shaders, often combining and reimagining prior ones in novel ways. The integration of AI into creative domains has gained momentum, with projects exploring the conversion of sketches into code and the auto-generation of HTML/CSS layouts from text descriptions. However, few systems support interactive, iterative, visually guided evolution, particularly in the context of shader evolution with real-time audio reactivity and user-in-the-loop mutation control. A recent development in AI-assisted programming is the concept of \textbf{vibe coding}, introduced by Andrej Karpathy in February 2025. Vibe coding refers to an AI-dependent programming technique where a person describes a problem in natural language prompts to an LLM tuned for coding, which then generates the corresponding software. This approach shifts the programmer's role from manual coding to guiding, testing, and refining the AI-generated code, making coding more accessible to those without extensive expertise.

Building upon these advancements, \textbf{AI Co-Artist} introduces a new design pattern that combines the generative capacity of LLMs with human aesthetic judgment in a loop of iterative refinement. In this system, the user acts as a creative curator, selecting visual outputs they find appealing, while the LLM evolves new shader code based on these preferences. Unlike traditional design tools, the user does not need to specify exact changes or understand the underlying codebase. Instead, they guide the system through aesthetic preference, making the process more accessible, exploratory, and engaging. This interactive evolution process mirrors natural creative cycles—brainstorm, prototype, refine—augmented by an AI assistant capable of producing diverse, structurally valid code variations. The result is a co-design paradigm where both the human and AI contribute meaningfully: the human sets the aesthetic trajectory, while the AI fills in technical detail and explores adjacent possibilities. Moreover, because LLMs understand a wide range of domains and programming languages, this framework can be easily extended beyond shaders, providing a glimpse into a broader future of AI-human collaboration in creative tasks.

\begin{figure*}[t]
\centering
\begin{tikzpicture}[
  every node/.style={rounded corners, font=\small},
  process/.style={rectangle, draw, fill=gray!15,
                  text width=3.6cm, minimum height=1.1cm, align=center},
  io/.style={rectangle, draw, fill=gray!5,
             text width=3.6cm, minimum height=1.1cm, align=center},
  arrow/.style={->, thick}
]

\node[io]      (user)    at (0, 0)    {User Interaction\\\small Selection, Evolution, Audio Upload};
\node[process] (audio)   at (6, 0)    {Audio Analysis\\\small MP3 $\rightarrow$ FFT $\rightarrow$ Feature};

\node[process] (validate) at (0, -2.4) {Shader Cleaning \&\\Sandbox Compilation};
\node[process] (llm)      at (6, -2.4) {LLM Shader Synthesis\\\small Mutation / Crossover};

\node[process] (render)  at (0, -4.8) {WebGL Renderer\\\small Multi-Canvas Population};
\node[io]      (display) at (6, -4.8) {Live Shader Display\\\small Fullscreen \& Grid View};

\draw[arrow] (user)  -- (audio);
\draw[arrow] (user)  -- (validate);
\draw[arrow] (audio) -- (llm);
\draw[arrow] (llm)   -- (validate);
\draw[arrow] (validate) -- (render);
\draw[arrow] (render) -- (display);

\end{tikzpicture}
\caption{System architecture of AI Co-Artist. Users interact through shader selection and audio upload. Audio is processed into a normalized scalar feature and passed to the WebGL renderer, while selected shaders are provided for LLM crossover or mutation. Generated shaders are compiled, validated and inserted back into the system.}
\label{fig:pipeline}
\end{figure*}
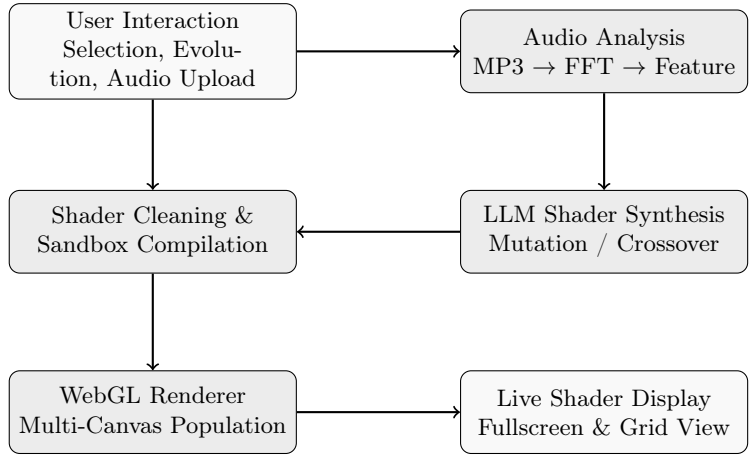

\section{System Architecture}
The system architecture of AI Co-Artist is composed of three major components: a WebGL-based real-time shader rendering engine, a client-side audio processing module using Tone.js, and a server-mediated GPT-4 interface for shader generation. Upon loading, the application presents a grid of shader canvases, each initialized with a randomly selected or pre-defined shader from a curated seed bank. Each canvas is independently animated, with time and audio analysis passed as uniforms. Audio files are uploaded by users in MP3 format. Using Tone.js, the system performs a real-time FFT decomposition of the signal, extracting frequency domain features. These features are normalized and provided to each shader via the \texttt{u\_audio} uniform, enabling dynamic, beat-reactive visual effects. User interaction occurs through selection. Users click on shaders they find aesthetically pleasing. Selected shaders are used as parents in either mutation (single parent) or crossover (multi-parent) generation operations. When triggered, the selected shader(s) are packaged into a structured natural language prompt and sent to the GPT-4 API. The LLM response is parsed to extract GLSL code, which is then compiled using WebGL. If the shader compiles successfully and renders without error, it replaces one of the non-selected shaders in the population. If compilation fails, up to five retry attempts are made with regenerated shaders. A notable design challenge is maintaining diversity while avoiding code bloat or performance degradation. To address this, shaders are constrained to a maximum code length, and rendering time per frame is capped. The system also supports full-screen previews and hides UI components during immersive exploration. The application runs in the browser without backend dependencies (except for LLM calls), ensuring privacy, low latency, and accessibility.

\subsection{WebGL-Based Real-Time Shader Rendering Engine}

AI Co-Artist is implemented as a browser-native, fully client-side system that integrates WebGL-based rendering, real-time audio analysis, and an LLM-driven shader evolution backend. The architecture is intentionally lightweight and portable, requiring no server-side computation beyond calls to the OpenAI API. The system transforms a user’s aesthetic preferences into an iterative evolutionary loop in which GLSL fragment shaders are dynamically generated, mutated, validated, rendered, and curated. Figure~\ref{fig:pipeline} illustrates the complete data and control flow among the major components. At the core of AI Co-Artist is a multi-canvas WebGL rendering engine that maintains a population of concurrently animated shaders. Each shader variant is rendered inside its own \texttt{<canvas>} element and is associated with an independent WebGL context, program, and rendering loop. The vertex shader is a minimal pass-through program that draws a full-screen quad, while the fragment shaders are supplied by the LLM or the curated base set. To ensure consistency across variants, the system automatically wraps each generated fragment shader with a standard interface that injects the uniforms \texttt{iTime}, \texttt{iResolution}, and \texttt{u\_audio}, followed by a unified entry point that forwards execution to \texttt{mainImage}. Each shader is compiled on-the-fly using WebGL, and invalid shaders are detected before they reach the user interface through an off-screen sandbox compilation stage.

\subsection{Client-Side Audio Processing and Feature Extraction}

Every shader variant runs in its own animation loop using \texttt{requestAnimationFrame}. During each frame, the system updates the uniform values according to the current canvas resolution, elapsed time since initialization, and audio-reactive feature described below. This design maintains high frame rates and smooth animations, even when managing a population of more than a dozen simultaneously evolving shaders. A long-press gesture enables fullscreen rendering for immersive exploration, while short clicks toggle selection, enabling user-driven curation. Audio reactivity is implemented using the Web Audio API and Tone.js. When the user uploads an MP3 file, the system constructs an analysis pipeline in which a \texttt{Tone.Player} streams audio through a 32-band FFT analyser. For each rendering frame, the analyser produces a frequency-domain vector, from which the system computes an aggregate audio energy value. This value is normalized into a continuous scalar in the range $\left[0,1\right]$, which is injected into every shader as the uniform \texttt{u\_audio}. Because this parameter modulates visual parameters without affecting temporal logic, shaders remain numerically stable while achieving expressive, beat-synchronous visual dynamics. The audio-reactive design also ensures that generated shaders do not require knowledge of the audio subsystem, keeping them portable and self-contained. Each shader receives the audio-reactive scalar $a_t = f(\text{FFT}_t)$ derived from FFT analysis, which modulates color, motion, or structural parameters. This term affects the rendered phenotype but not the underlying genotype (shader code). Thus, the mapping $\text{phenotype}(S, a_t)$ is dynamic and time-varying, influencing user evaluation and hence the evolutionary trajectory.

\subsection{GPT-4–Mediated Shader Generation and Evolution Module}

AI Co-Artist maintains a population of 14 shader variants, each represented by a canvas, shader program, and metadata describing its selection state and fullscreen mode. Selection is performed solely through short-click interactions, thereby making aesthetic curation the dominant mode of user engagement. The \texttt{Evolve} button becomes active once a non-empty subset of variants is selected. When the user invokes evolution, the system determines whether to apply mutation (single parent) or crossover followed by mutation (multiple parents). Unselected variants are replaced with newly generated offspring, while selected shaders persist as elite individuals, enabling directional exploration of the visual design space. To ensure robustness, newly created shader code is cleaned, validated syntactically, and compiled in a sandbox before becoming part of the population. The system retries generation up to a configurable number of attempts if compilation fails. This prevents unusable shaders from entering the population and sustains a smooth interactive exploration experience. The evolutionary process is powered by GPT-4 through structured prompts. For each evolution event, the system constructs either a mutation or crossover prompt, embedding one or more parent shaders inside a natural language description that requests structural preservation, artistic coherence, and syntactically valid GLSL code. The LLM is queried with a moderately high sampling temperature to encourage aesthetic diversity, and the response is parsed to extract raw shader code without explanations or formatting. The resulting fragment shader is checked for compilation correctness before being admitted into the population. This LLM-mediated pipeline functions as a semantic mutation operator, enabling complex transformations that exceed traditional stochastic mutations while maintaining validity and expressivity.

\section{Methodology}
AI Co-Artist is designed as a fully interactive, web-based system that combines WebGL shader rendering, real-time audio reactivity, and LLM-powered shader synthesis into an evolutionary art creation experience. The methodology consists of five major components: (1) initialization and population of shader variants, (2) real-time audio feature extraction, (3) user interaction and selection, (4) shader evolution via LLM-driven crossover and mutation, and (5) shader replacement and iteration. Upon loading, the system initializes a shader population with a predefined number of GLSL fragment shaders, either from a curated base set or from generated mutations. The base shaders are designed to be visually diverse and audio-reactive, leveraging the uniform variable \(\texttt{u\_audio}\) derived from the audio input. If the total number of initial shaders (defined as \texttt{POPULATION\_SIZE}) exceeds the number of base shaders, additional shaders are generated via mutation using an OpenAI LLM (e.g., GPT-4). Each shader is rendered in an independent WebGL canvas element using a common vertex shader and a dynamically compiled fragment shader. All canvases are embedded in a responsive grid layout for comparative visual exploration. For each new shader variant, the system dynamically compiles the GLSL code using WebGL and updates the rendering canvas. All shaders are sandboxed per canvas, ensuring isolation and parallel rendering. The canvas can be toggled to full-screen mode through a long press (or hold gesture), enabling immersive exploration.
Failed shaders are excluded from rendering to maintain visual consistency. Successfully compiled shaders update the respective slot, effectively evolving the population. All generated shaders undergo syntax validation and WebGL compilation checks. If a generated shader fails, the system retries up to a maximum number of attempts (\texttt{MAX\_ATTEMPTS}) before falling back to a base shader.

\subsection{Interactive Evolution Loop}

AI Co-Artist implements an evolutionary process in which users guide the selection process while large language models (LLMs) perform semantic mutation and crossover to generate new shaders. In this section, we present a formal model of this interaction, capturing the human--AI feedback loop, population dynamics, and LLM-based transformation operators. This formalization provides a foundation for analyzing stability, diversity, and convergence properties of the system. Let $\mathcal{P}_t = \{S_t^{(1)}, S_t^{(2)}, \dots, S_t^{(N)}\}$ denote the shader population at iteration $t$, where each $S_t^{(i)}$ is a GLSL fragment shader program represented as a string. In AI Co-Artist, $N = 14$. Each program is compiled and rendered independently within the WebGL engine and evaluated visually by the user. Because shaders share a fixed interface (uniforms, main entry point), they occupy a constrained but expressive program space $S \in \mathcal{S} \subseteq \text{GLSL}$ where $\mathcal{S}$ is the subspace of fragment shaders compatible with the system’s rendering pipeline. At each iteration, the user selects an arbitrary non-empty subset $\mathcal{E}_t \subseteq \mathcal{P}_t$ representing elite individuals whose visual characteristics the user wishes to preserve. Selection is therefore non-numeric, qualitative, and subjective, reflecting aesthetic judgment rather than fitness evaluated by a predefined objective. Unselected shaders constitute the replaceable set $\mathcal{R}_t = \mathcal{P}_t \setminus \mathcal{E}_t$. The complete interaction cycle is therefore:
\[
\underbrace{\mathcal{P}_t \xrightarrow{\text{render}} \text{visuals}}_{\text{WebGL}}
\;\xrightarrow{\text{user selection}}\;
\mathcal{E}_t
\;\xrightarrow{\text{LLM operators}}\;
\mathcal{P}_{t+1}
\]

This defines a mixed-initiative evolutionary system in which humans provide qualitative selection pressure and the LLM executes semantic variation. Unlike classical evolutionary algorithms, where mutation and crossover are stochastic and local, the LLM enables globally meaningful transformations informed by learned priors from code and artistic corpora. The result is an evolutionary process that is expressive, user-driven, and semantically rich, enabling high-level creative exploration while preserving shader validity across generations. The evolution cycle is fully user-driven. The user begins by selecting one or more shaders that they find visually appealing. This selection triggers the evolutionary logic: (1) if one shader is selected, mutation is applied, (2) if multiple shaders are selected, crossover is applied first to synthesize a parent, followed by mutation to generate variants, and (3) non-selected shaders are then replaced by these new variants. This loop allows continuous aesthetic refinement through intuitive curation, enabling an active participation in the process. Each iteration allows the user to progressively guide the population toward a desired aesthetic space.

\subsection{LLM as Evolution Operators}

The core evolution mechanism is driven by GPT-4 (or equivalent) through structured prompt engineering. Depending on user selection, the system applies one of two evolutionary strategies:
\begin{itemize}
\item \textbf{Mutation}: When a single shader is selected, it is passed to the LLM along with a prompt to generate a creatively modified but structurally similar variant. This promotes stylistic exploration while maintaining coherence.
\begin{quote}
\textit{You are a world-recognized artist and shader programmer. Create an innovative and improved variant of the provided GLSL fragment shader code. Preserve its structure while creatively enhancing visual dynamics. Provide only the GLSL shader code without explanations.}
\end{quote}
\item \textbf{Crossover}: When multiple shaders are selected, their source codes are included in a prompt asking the LLM to synthesize a coherent hybrid that blends stylistic and structural attributes. The result serves as a new base for further mutation.
\begin{quote}
\textit{You are a renowned shader artist. Create a cohesive hybrid shader by blending the visual attributes of the following GLSL shaders. Return only the resulting GLSL shader code without any explanations.}
\end{quote}
\end{itemize}

Given the selected set $\mathcal{E}_t$, the system determines whether to apply mutation or crossover. For a single selected shader $S$, mutation produces a semantic variant $M_{\theta}(S) = S'$, where $M_{\theta}$ is the LLM transformation parameterized by model weights $\theta$. Mutation introduces stylistic and structural changes while preserving functional integrity. For a set of parents $\{S_1, S_2, \dots, S_k\}$, crossover produces a blended hybrid $C_{\theta}(S_1, \dots, S_k) = S^\star$, where $S^\star$ inherits conceptual patterns from multiple parents. The hybrid is typically further refined by mutation $S' = M_{\theta}(S^\star)$. The LLM thus serves as a high-level, knowledge-aware recombination engine. New shaders replace only unselected individuals. Formally, at each iteration:
\[
\mathcal{P}_{t+1}
= \mathcal{E}_t
\cup
\{ M_{\theta}(S^\star_j) \mid j = 1, \dots, |\mathcal{R}_t| \},
\]
where $S^\star_j$ is either a mutated elite shader (single-parent case) or a crossover-derived hybrid (multi-parent case). This ensures that user-selected programs persist across generations while the remainder of the population explores new aesthetic directions. The success of the system hinges on carefully crafted prompts that steer the LLM toward generating visually diverse, syntactically correct, and semantically meaningful shaders. Prompts are designed to be minimal yet expressive, invoking the LLM’s creativity without overwhelming its token window:

The \textbf{mutation prompt} takes a single shader as input and asks the model to improve upon it. It emphasizes creativity, aesthetic refinement, and structural preservation. The intent is to nudge the model to explore adjacent variants of a known good shader, adding new elements or modifying dynamics without straying too far. The \textbf{crossover prompt} is more exploratory. It provides multiple shaders and asks the model to create a coherent hybrid. This invites the LLM to blend stylistic and structural elements, leading to more novel and diverse outputs. These prompts are tested across dozens of iterations and continuously refined. We found that explicitly invoking the model’s artistic role ("world-recognized artist") improved the coherence and boldness of outputs. Encouraging structure preservation reduced the rate of compilation errors.

\section{Discussion}

AI Co-Artist employs a novel form of semantic evolution in which mutation and crossover are delegated to a large language model rather than implemented through classical stochastic operators. This design choice yields transformations that are syntactically valid, aesthetically meaningful, and context-aware in ways that conventional symbolic mutation often fails to achieve. In this section, we examine the functional behavior, structural characteristics, and emergent properties of the two LLM-driven operators, drawing parallels to evolutionary computation while highlighting their distinct advantages.

\subsection{Semantic Mutation via LLM Transformation}

Traditional mutation operators in evolutionary art systems perturb numerical constants, insert noise patterns, or randomly modify syntactic structures. However, such operations often produce invalid GLSL shaders or degenerate outputs. By contrast, AI Co-Artist formalizes mutation as a \emph{semantic rewriting} operation performed by the LLM. Given a parent shader $S$, the mutation operator $M$ produces a variant $S'$ by generating a meaning-preserving but visually distinct reformulation:
\[
S' = M_{\theta}(S),
\]
where $M_{\theta}$ denotes the LLM transformation under model parameters $\theta$.

Empirically, the LLM exhibits several characteristic mutation modes:

\begin{itemize}
    \item \textbf{Parameter Mutations.} The model frequently alters numeric constants, oscillation frequencies, color weights, or smoothing factors. These adjustments preserve the underlying structure while creating perceptible aesthetic variation.
    
    \item \textbf{Spatial Structure Mutations.} The LLM modifies coordinate warping expressions, distance estimators, or rotation matrices, thereby producing new spatial patterns such as ripples, spirals, or domain-warped textures.
    
    \item \textbf{Color Mapping Mutations.} A common mode involves introducing new gradient mappings, nonlinear color transforms, or palette functions. These changes yield substantial variation while retaining compatibility with the core rendering logic.
    
    \item \textbf{Temporal and Audio Reactivity Enhancements.} The model occasionally introduces modulation terms involving \texttt{iTime} or \texttt{u\_audio}, producing richer animations. Because the prompts explicitly encourage ``enhancing visual dynamics,'' such behavior is frequent.
\end{itemize}

Unlike random perturbations, these mutations are coherent and intentional, guided by the LLM's internal representation of artistic structure. This makes each mutation ``high-level'' in the sense that it preserves semantic intent even when rewriting code substantially.

\subsection{Crossover as Semantic Blending}

In classical evolutionary computation, crossover combines substrings or parse trees from two parent genomes. However, fragment shaders do not lend themselves naturally to syntactic crossover: swapping arbitrary code blocks often breaks variable scope, function dependencies, or numerical stability. The LLM overcomes this by performing crossover as a \emph{conceptual blending} operation:
\[
S_{\text{hybrid}} = C_{\theta}(S_1, S_2, \ldots, S_n),
\]
where $C_{\theta}$ instructs the model to ``blend'' or ``combine'' visual characteristics, enabling it to merge conceptual elements across parents.

Analysis of generated hybrids reveals several consistent crossover behaviors:

\begin{itemize}
    \item \textbf{Palette Transfer.} The LLM often inherits the color palette from one shader while using the spatial structure of another. For example, a fractal pattern from $S_1$ may be recolored using the neon palette from $S_2$.
    
    \item \textbf{Structural Fusion.} The model combines different structural motifs by integrating their coordinate transformations, noise functions, or domain-warping techniques. This results in visually complex hybrids that often exceed the expressive capacity of a single parent.
    
    \item \textbf{Behavioral Blending.} The temporal or audio-reactive behaviors of multiple parents may be fused, e.g., adopting the beat-driven pulse from one parent while retaining rotation dynamics from another.
    
    \item \textbf{Function Merging.} The LLM occasionally merges helper functions (e.g., noise, fbm, or pattern functions) into a unified shader, rewriting variable names as necessary to preserve correctness.
\end{itemize}

This semantic crossover is particularly powerful because it reflects the LLM’s understanding of the functional and visual intent of each codebase, rather than blindly recombining syntax. The resulting hybrids are typically well-formed and visually rich, avoiding the brittleness associated with structural crossover operators in DSL-based genetic programming.

\subsection{Emergent Properties of LLM-Driven Evolution}

LLM-based mutation and crossover introduce several emergent phenomena that do not arise in classical evolutionary computation:

\begin{itemize}
    \item \textbf{High-Level Exploration.} The space of possible mutations spans high-level artistic concepts, such as symmetry, contrast, or motion, rather than simple numeric perturbations.
    
    \item \textbf{Implicit Domain Knowledge.} The model frequently employs advanced shader patterns—domain warping, ray-marching, layered noise—without being explicitly programmed to do so.
    
    \item \textbf{Robustness to Structural Variations.} Because the LLM preserves interfaces and rewrites code holistically, shaders remain compilable across generations, ensuring evolutionary stability.
    
    \item \textbf{Non-Local Variation.} Mutations may involve coordinated changes across multiple lines of code, producing consistent behavior changes that would be improbable through random mutation.
\end{itemize}

These behaviors position the LLM not merely as a mutation engine, but as a domain-savvy creative collaborator capable of producing meaningful aesthetic jumps while respecting execution constraints.

\subsection{Comparison to Classical Evolutionary Operators}

Table~\ref{tab:comparison} summarizes the contrast between LLM-driven and classical evolutionary operators. The LLM approach offers semantic coherence, high-level creativity, and structural reliability, at the cost of increased computational overhead and reliance on external model inference. In summary, the mutation and crossover operators in AI Co-Artist leverage LLMs to perform meaning-aware transformations that emulate human-like shader editing. This enables a richer and more reliable evolutionary process than traditional genetic operators, and plays a central role in the system's ability to support open-ended creative exploration.

\begin{table}[h]
\centering
\caption{Comparison of classical and LLM-driven evolutionary operators.}
\begin{tabular}{p{3cm} p{4cm} p{4cm}}
\toprule
\textbf{Aspect} & \textbf{Classical Mutation/Crossover} & \textbf{LLM-Driven Mutation/Crossover} \\
\midrule
Granularity & Local numeric/syntactic perturbation & Global semantic rewriting \\
Validity & Often fragile, frequent syntax errors & High validity due to holistic rewriting \\
Creativity & Low to moderate & High, guided by learned artistic priors \\
Domain Knowledge & None & Implicit from training data \\
Computational Cost & Low & Higher (LLM inference) \\
User Control & Indirect & Direct via prompt and selection \\
\bottomrule
\end{tabular}
\label{tab:comparison}
\end{table}

\section{User Experience Study}
AI Co-Artist is designed around the principle of human-in-the-loop creativity, allowing users to guide shader evolution solely through visual preferences. The system minimizes technical friction while maximizing aesthetic agency. Upon launching the interface, the user is greeted with a grid of audio-reactive shaders rendered in real-time. Each shader is displayed in its canvas, and all are animated simultaneously. Users can upload their own MP3 files to drive the animations. Once an audio file is uploaded, all shaders are synchronized to the audio's beat and frequency structure. Each canvas is interactive:
\begin{itemize}
\item \textbf{Short click or tap}: Selects or deselects the shader. A white border highlights selected shaders.
\item \textbf{Long press (1.5s)}: Triggers fullscreen preview of the shader, hiding UI elements for an immersive experience. Releasing the press exits fullscreen.
\end{itemize}
These interactions support rapid browsing, comparative evaluation, and focused inspection of visual effects. Once the user selects one or more shaders, the \textbf{\texttt{Evolve}} button is enabled. Upon clicking it:
\begin{enumerate}
\item The selected shaders are used to generate a new parent (via crossover and mutation).
\item The system then applies mutations to this parent shader.
\item All non-selected shaders in the population are replaced by new mutated variants.
\end{enumerate}
This allows the user to retain visually compelling outputs while evolving the rest of the population. Selected shaders can be downloaded via the \texttt{Download} button. This saves all selected GLSL codes into a plain text file, allowing users to reuse, remix, or publish their creations outside the platform.
This interaction model enables both novices and professionals to steer complex shader evolution using nothing but aesthetic judgment, eliminating the need for code manipulation while maintaining rich creative control.

To evaluate the effectiveness and usability of AI Co-Artist, we conducted a structured user study involving 50 participants recruited from digital art communities, shader coding forums, and university courses. Participants were divided into two balanced cohorts: shader novices (no coding experience, n=30) and shader experts (moderate to advanced GLSL proficiency, n=20). Each participant was asked to complete two tasks in randomized order: (1) create as many visually interesting audio-reactive shaders as possible in 25 minutes using AI Co-Artist, and (2) perform the same task using Shadertoy.com without AI assistance. We measured quantitative metrics, including: (1) Average number of completed shaders, (2) Time to first visually pleasing result, (3) Compilation error rate, and User satisfaction (Likert scale 1--5). Novice users created an average of 4.2 shaders with AI Co-Artist vs 0.6 with Shadertoy. Experts created 6.8 vs 2.9. Time to first viable output was reduced by over 60\% for both groups. Satisfaction averaged 4.7/5 (AI Co-Artist) vs 2.8/5 (Shadertoy). Compilation errors occurred in less than 3\% of generations after retries. Qualitative feedback indicated that users appreciated the "collaborative feeling" with the AI. Novices described the system as "magical" and "fun," while experts saw it as an ideation partner. Many participants requested extensions into other creative domains.

\section{Conclusion}
AI Co-Artist demonstrates how LLMs can be effectively integrated into creative workflows through interactive evolution. By combining real-time shader rendering, audio reactivity, and LLM-generated code, the system provides a novel platform for visual exploration. Users serve as curators, while the LLM serves as an ideation engine. Our user study confirms that AI Co-Artist improves productivity, reduces learning curves, and enhances satisfaction across user types. It stands as a compelling case for human-AI collaboration in digital art, and offers a template for similar systems in adjacent domains. AI Co-Artist lowers the barrier to shader art, democratizing access to a domain that was previously limited to technically skilled individuals. By providing a co-creative AI agent, it enables a broader audience—including musicians, students, and designers—to explore procedural visual art without programming expertise. We plan to extend AI Co-Artist beyond shader design to encompass broader creative domains, such as automated generation of responsive web layouts (HTML/CSS), architectural concept sketches (via procedural modeling), and product designs using LLMs trained on domain-specific design languages. Lastly, we aim to incorporate multimodal vision-language models as autonomous judges in the selection loop. Instead of relying solely on human input, the system could use pretrained visual models to evaluate generated shaders against aesthetic criteria or user-defined moodboards. This would enable fully autonomous aesthetic evolution loops.


\begin{thebibliography}{8}

\bibitem{takagi2001interactive}
Takagi, H.: Interactive evolutionary computation: Fusion of the capabilities of EC optimization and human evaluation. \textit{Proceedings of the IEEE} \textbf{89}(9), 1275--1296 (2001)

\bibitem{lewis2008genetic}
Lewis, M.: Evolutionary visual art and design. In: The Art of Artificial Evolution, pp. 3--37. Springer, Heidelberg (2008)

\bibitem{stanley2002evolving}
Stanley, K.O., Miikkulainen, R.: Evolving neural networks through augmenting topologies. \textit{Evolutionary Computation} \textbf{10}(2), 99--127 (2002)

\bibitem{secretan2011picbreeder}
Secretan, J., Beato, N., D'Ambrosio, D.B., Rodriguez, A., Campbell, J., Stanley, K.O.: Picbreeder: A case study in collaborative evolutionary exploration of design space. \textit{Evolutionary Computation} \textbf{19}(3), 373--403. MIT Press (2011)

\bibitem{biles1994genjam}
Biles, J.A.: GenJam: A genetic algorithm for generating jazz solos. In: Proceedings of the International Computer Music Conference, pp. 131--137 (1994)

\bibitem{chen2021codex}
Chen, M., Tworek, J., Jun, H., Yuan, Q., Ponde, H., Kaplan, J., Edwards, H., Burda, Y., Joseph, N., Brockman, G., et al.: Evaluating large language models trained on code. \textit{arXiv preprint arXiv:2107.03374} (2021)

\bibitem{wang2023codellama}
Wang, L., Zhang, Z., Chowdhery, A., Li, X., Lepikhin, D., et al.: Code Llama: Open foundation models for code. \textit{arXiv preprint arXiv:2308.12950} (2023)

\bibitem{chen2023shader}
Chen, Y., Lin, Y., Guo, J., Huang, W.: Prompting large language models for shader programming. In: Proceedings of the 2023 ACM Symposium on User Interface Software and Technology. ACM, New York (2023)

\end{thebibliography}
\end{document}